\pdfoutput=1
\documentclass[11pt,a4paper]{article}
\usepackage[hyperref]{emnlp2020}
\usepackage{times}
\usepackage{latexsym}

\usepackage{graphicx}
\usepackage{url}
\usepackage{multirow}
\usepackage{subcaption}
\usepackage{amsmath,bm,amsfonts,dsfont,bbm}
\usepackage{paralist}
\usepackage{hyperref}
\usepackage{multicol}
\usepackage{color}
\usepackage{booktabs}
\usepackage[inline]{enumitem}
\usepackage{xspace}
\usepackage{mdframed}
\usepackage[T1]{fontenc}
\usepackage[utf8]{inputenc}

\usepackage[activate={true,nocompatibility}, kerning=true,spacing=true, stretch=5,shrink=25]{microtype}

\aclfinalcopy %
\def\aclpaperid{1518} %

\newcommand\Mark[1]{\textsuperscript#1}

\author{Shubham Toshniwal\Mark{1},
Sam Wiseman\Mark{1},
Allyson Ettinger\Mark{2},
Karen Livescu\Mark{1},
Kevin Gimpel\Mark{1}
\\
\Mark{1}Toyota Technological Institute at Chicago\\
\Mark{2}Department of Linguistics, University of Chicago\\[0.5em]
\small{\texttt{\{shtoshni, swiseman, klivescu, kgimpel\}@ttic.edu, aettinger@uchicago.edu}}\\
}

\definecolor{dkgreen}{RGB}{0,130,0}
\definecolor{aqua}{rgb}{0.0, 0.4, 1.0}

\newcommand{\kevin}[1]{\textcolor{dkgreen}{$_{\mathit{KG}}$[#1]}}
\newcommand{\karen}[1]{\textcolor{blue}{$_{KL}$[#1]}}
\newcommand{\ake}[1]{\textcolor{red}{$_{AE}$[#1]}}

\newcommand{\st}[1]{\textcolor{brown}{$_{ST}$[#1]}}

\renewcommand{\kevin}[1]{}
\renewcommand{\karen}[1]{}
\renewcommand{\ake}[1]{}
\renewcommand{\st}[1]{}

\newcommand{\unbounded}{U-MEM\xspace}
\newcommand{\learned}{LB-MEM\xspace}
\newcommand{\lru}{RB-MEM\xspace}

\title{Learning to Ignore: Long Document Coreference \\ with Bounded Memory Neural Networks} %

\date{}

\begin{document}
\maketitle
\begin{abstract}
Long document coreference resolution remains a challenging task	 due to the large memory and runtime requirements of current models.
Recent work doing incremental coreference resolution using just the global representation of entities shows practical benefits but requires keeping all entities in memory, which can be impractical for long documents. %
We argue that keeping all entities in memory is unnecessary, and we propose a memory-augmented neural network that tracks only a small bounded number of entities at a time, thus guaranteeing a linear runtime in length of document.
We show that (a) the model remains competitive with models with high memory and computational requirements on OntoNotes and LitBank, and (b) the model learns an efficient memory management strategy easily outperforming a rule-based strategy.

\end{abstract}

\section{Introduction}
Long document coreference resolution poses runtime and memory challenges.
Current best models %
for coreference resolution have large memory requirements and quadratic runtime in the document length~\citep{joshi-etal-2019-bert, wu2019coreference}, making them impractical for long documents. %

Recent work revisiting the entity-mention paradigm~\citep{luo-etal-2004-mention, websterC14}, which seeks to maintain explicit representations only of entities, rather than all their constituent mentions, has shown practical benefits for memory while being competitive with state-of-the-art models~\citep{xia2020revisiting}.
In particular, unlike other approaches to coreference resolution which maintain representations of both mentions \textit{and} their corresponding entity clusters~\citep{rahman2011narrowing, stoyanov-eisner-2012-easy, clark-manning-2015-entity, wiseman-etal-2016-learning,lee-etal-2018-higher}
, the entity-mention paradigm stores representations only of the entity clusters, which are updated incrementally as coreference predictions are made.
While such an approach requires less memory than those that additionally store mention representations, the number of entities can be impractically large when processing long documents, making the storing of all entity representations problematic.

Is it necessary to maintain an unbounded number of mentions or entities?  Psycholinguistic evidence suggests it is not, as human language processing is incremental \citep{Tanenhaus1632, keller2010cognitively} and has limited working memory~\citep{baddeley1986}.
In practice, we find that most entities have a small spread (number of tokens from first to last mention of an entity), and thus do not need to be kept persistently in memory.
This observation suggests that tracking a limited, small number of entities at any time can resolve the computational %
 issues, albeit at a potential accuracy tradeoff.

Previous work on bounded memory models for coreference resolution has shown potential, but has been tested only on short documents  %
\citep{liu2019referential, toshniwal2020petra}. %
Moreover, this previous work makes token-level predictions while standard coreference datasets have span-level annotations.  %
We propose a bounded memory model that performs quasi-online coreference resolution,\footnote{``Quasi-online'' because document encoding uses bi-directional transformers with access to future tokens.}
 and test it on LitBank~\cite{bamman2019annotated} and OntoNotes~\citep{pradhan2012conll}.
The model is trained to manage its limited memory by predicting whether to ``forget" an entity already being tracked in exchange for a new (currently untracked) entity. %
Our empirical results show that: (a) the model is competitive with an unbounded memory version, and (b) the model's learned memory management outperforms a strong rule-based baseline.%
\footnote{\url{https://github.com/shtoshni92/long-doc-coref}}

\section{Entity Spread and Active Entities}
\begin{table}[t]
    \centering{
    \small{
    \caption{Max.\ Total Entity Count vs.\ Max.\ Active Entity Count.}
    \label{sec:tab_active_entities}

    \begin{tabular}{lcc}
    \toprule
        & LitBank & OntoNotes \\\midrule
    Max.\ Total Entity Count &  199 & 94 \\
    Max.\ Active Entity Count & $\phantom{1}$18 & 24 \\\bottomrule
    \end{tabular}
    }
    }
\end{table}

\label{sec:active}
Given input document $\mathcal{D}$, let $(x_n)_{n=1}^N$ represent the $N$ mention spans corresponding to $M$ underlying entities $(e_m)_{m=1}^M$. %
Let $\textrm{START}(x_i)$ and $\textrm{END}(x_i)$ denote the start and end token indices of the mention span $x_i$ in document $\mathcal{D}$.
Let $\textrm{ENT}(x_i)$ %
 denote the entity of which $x_i$ is a mention.
Given this notation we next define the following concepts.

\paragraph{Entity Spread} Entity spread denotes the interval of token indices from the first mention to the last mention of an entity. %
The entity spread $\textrm{ES}(e)$ of entity $e$ is given by: %
$$\textrm{ES}(e) = [\min_{\textrm{ENT}(x) = e}\textrm{START}(x), \max_{\textrm{ENT}(x) = e}\textrm{END}(x)]$$

\paragraph{Active Entity Count}
Active entity count $\textrm{AE}(t)$ at token index $t$  denotes the number of unique entities whose spread covers the token $t$, i.e.,
$\textrm{AE}(t) = |\{e \;|\; t \in \textrm{ES}(e) \}|$.

\paragraph{Maximum Active Entity Count}
Maximum active entity count $\textrm{MAE}(\mathcal{D})$ for a document $\mathcal{D}$ denotes the maximum number of active entities at any token index in $\mathcal{D}$, i.e.,
$\textrm{MAE}(\mathcal{D}) = \max_{t \in [|\mathcal{D}|]} \textrm{AE}(t)$. 
This measure can be simply extended to a corpus $\mathcal{C}$ as: $\textrm{MAE}(\mathcal{C}) = \max_{\mathcal{D} \in \mathcal{C}} \textrm{MAE}(\mathcal{D})$. %

Table~\ref{sec:tab_active_entities} shows the MAE and the maximum total entity count in a single document,
 for LitBank and OntoNotes. %
For both datasets the maximum active entity count is much smaller than the maximum total entity count.
Thus, rather than keeping all the entities in memory at all times, models can in principle simply focus on the far fewer active entities at any given time.

\section{Model}
Based on the preceding finding, we will next describe models that require tracking only a small, bounded number of entities at any time.

To make coreference predictions for a document, we first encode the document and propose candidate mentions.
The proposed mentions are then processed sequentially and are either: (a) added to an existing entity cluster, (b) added to a new cluster, %
 (c) ignored due to limited memory capacity (for bounded memory models), or (d) ignored as an invalid mention.

\paragraph{Document Encoding}
is done using the SpanBERT\textsubscript{LARGE} model finetuned for OntoNotes and released as part of the coreference model of \citet{joshi2020span}.
We don't further finetune the SpanBERT model.
To encode long documents, we segment the document using the \emph{independent} and \emph{overlap} strategies described in \citet{joshi-etal-2019-bert}.\footnote{We modify the \emph{overlap} segmentation to respect sentence boundary or token boundary when possible.}
In \emph{overlap} segmentation, for a token present in overlapping BERT windows, the token's representation is taken from the BERT window with the most neighboring tokens of the concerned token.
For both datasets we find that \emph{overlap} slightly outperforms \emph{independent}.

\paragraph{Mention Proposal}
\label{sec:ment_proposal}
Given the encoded document, we next predict the top-scoring mentions which are to be clustered.
The goal of this step is to have high recall, and we follow previous work to threshold the number of spans chosen~\citep{lee-etal-2017-end}.
Given a document $\mathcal{D}$, we choose $0.3 \times |\mathcal{D}|$ top spans for LitBank, and  $0.4 \times |\mathcal{D}|$ for OntoNotes.

We pretrain the mention proposal model before training the mention proposal and mention clustering pipeline end-to-end, as done by \citet{wu2019coreference}.
The reason is that without pretraining, most of the mentions proposed by the mention proposal model would be invalid mentions, i.e., spans that are not mentions, which would not provide any training signal to the mention clustering stage.

\paragraph{Mention Clustering}
Let $(x_i)_{i=1}^K$ represent the top-$K$ candidate mention spans from the mention proposal step and let $s_m(x_i)$ represent the mention score for span $x_i$, which indicates how likely it is that a span constitutes a mention.
Assume that the mentions are already ordered based on their position in the document and are processed sequentially in that order.\footnote{Specifically, they are ordered based on START($\cdot$) index with ties broken using END($\cdot$).}
Let $E = (e_m)_{m=1}^M$ represent the $M$ entities currently being tracked by the model (initially $M = 0$).
For ease of discussion, we will overload the terms $x_i$ and $e_j$ to also correspond to their respective representations.

In the \emph{first} step, the model decides whether the span $x_i$ refers to any of the entities in $E$ as follows:
\begin{align*}
s_c(x_i, e_j) &\!=\! f_c([x_i; e_j; x_i \odot e_j; g(x_i, e_j)]) \!+\! s_m(x_i)\\
s_c^{\mathit{top}} &\!=\! \max_{j=1 \dotsc M} s_c(x_i, e_j)\\
e^{\mathit{top}} &\!=\! \underset{{j=1 \dotsc M}}{\arg\max}\; s_c(x_i, e_j)
\end{align*} %
where $\odot$ represents the element-wise product, and $f_c(\cdot)$ corresponds to a learned feedforward neural network.
The term $g(x_i, e_j)$ correponds to a concatenation of feature embeddings that includes embeddings for (a) number of mentions in $e_j$, (b) number of mentions between $x_i$ and last mention of $e_j$, (c) last mention decision, and (d) document genre (only for OntoNotes). %

Now if $s_c^{\mathit{top}} > 0$ then $x_i$ is considered to refer to $e^{\mathit{top}}$, and $e^{\mathit{top}}$ is updated accordingly.\footnote{We use weighted averaging where the weight for $e^{\mathit{top}}$ corresponds to the number of previous mentions seen for  $e^{\mathit{top}}$.}
Otherwise, $x_i$ does not refer to any entity in $E$ and a \emph{second} step is executed, which will depend on the choice of memory architecture.
We test three
 memory architectures, described below.

\begin{enumerate}[wide, labelwidth=!, labelindent=0pt]
    \item \textbf{Unbounded Memory (\unbounded)}: If $s_m(x_i) > 0$ then we create a new entity $e_{M + 1} = x_i$ and append it to $E$.
    Otherwise the mention is ignored as invalid, i.e., it doesn't correspond to an entity.
    Ignoring invalid mentions is important for datasets such as LitBank where singletons are explicitly annotated and used for evaluation.
    For OntoNotes, where singletons are not annotated and ignored for evaluation, we also consider a variant U-MEM* which appends all non-coreferent mentions, as done in \citet{xia2020revisiting}.

    \item \textbf{Bounded Memory}: Suppose the model has a capacity of tracking $C$ entities at a time.
    If $C > M$, i.e., the memory capacity has not been fully utilized, then the model behaves like U-MEM. %
    Otherwise, the bounded memory models must decide between: (a) evicting an entity already being tracked, (b) ignoring $x_i$ due to limited capacity, and (c) ignoring the mention as invalid.
    We test two bounded memory variants that are described below.

    \begin{enumerate}[wide, labelwidth=!, labelindent=0pt]
        \item \textbf{Learned Bounded Memory (\learned)}:
        The proposed \learned architecture tries to predict a score $f_r(.)$ corresponding to the anticipated number of remaining mentions for any entity or mention, and compares it against the mention score $s_m(x_i)$ as follows:
        \begin{align*}
            &d = \arg\min [f_r(e_1), \dotsc, f_r(e_M), f_r(x_i), s_m(x_i)]
        \end{align*}
        where $f_r(\cdot)$ is a learned feedforward neural network. %
        If $1 \le d \le M$ then then the model evicts the previous entity $e_d$ and reinitialize it to $x_i$.
        Otherwise if $d = M + 1$ then the model ignores $x_i$ due to limited capacity.
        Finally if $d = M + 2$ then the model predicts the mention to be invalid.
        \item \textbf{Rule-based Bounded Memory (\lru)}
        The Least Recently Used (LRU) principle is a popular choice among memory models~\citep{rae2016scaling, santoro2016one}.
        While \learned  considers all potential entities for eviction, with \lru  this choice is restricted to just the LRU entity, i.e., the entity whose mention was least recently seen.
        The rest of the steps are similar to the \learned model.
    \end{enumerate}
\end{enumerate}

\paragraph{Training}
All the models are trained using teacher forcing.
The ground truth decisions for bounded memory models are chosen to maximize the number of mentions tracked by the model (details in Appendix~\ref{sec:app_gt}).
Finally, the training loss is calculated via the addition of the cross-entropy losses for the two steps of mention clustering.

\section{Experimental Setup}

\label{sec:res_litbank}
\begin{table}[t]

\centering{
\small{
\caption{Results for LitBank (CoNLL F1).}
\label{tab:res_litbank}

\begin{tabular}{lcc}
\toprule
Model & Dev F1 & Test F1 \\\midrule
\unbounded & 77.1 & 76.5 \\
\learned &  & \\
\hspace{0.1in} 5 cells & 71.9 &  70.3 \\
\hspace{0.1in} 10 cells & 75.0 & 74.7 \\
\hspace{0.1in} 20 cells & 75.7 & 75.1 \\
\lru & & \\
\hspace{0.1in} 5 cells & 58.5 & 57.8 \\
\hspace{0.1in} 10 cells & 69.9 & 69.0 \\
\hspace{0.1in} 20 cells & 75.3 & 74.4 \\\midrule
\citet{bamman2019annotated} & - & 68.1 \\
\bottomrule
\end{tabular}
}
}
\end{table}

\subsection{Datasets}

\paragraph{LitBank} $\!\!\!\!\!$ is a recent coreference dataset for literary texts~\citep{bamman2019annotated}.
The dataset consists of prefixes of 100 novels with an average length of 2100 words.
Singletons are marked and used for evaluation.  %
Evaluation is done via 10-fold cross-validation over 80/10/10 splits.\footnote{\url{https://github.com/dbamman/lrec2020-coref/tree/master/data}}

\paragraph{OntoNotes} $\!\!\!\!$ consists of 2802/343/348 documents in the train/development/test splits, respectively \cite{pradhan2012conll}.
The documents span 7 genres and have an average length of 463 words.
Singletons are not marked in the dataset.

\subsection{Hyperparameters}
\label{sec:hyperparam}

Document encoding is done using the SpanBERT\textsubscript{LARGE} model of \citet{joshi2020span} which was finetuned for OntoNotes.\footnote{
From the original models, we stripped out just the SpanBERT part which is available at \url{https://huggingface.co/shtoshni/spanbert_coreference_large}
}  %
The SpanBERT model is not further finetuned.
The other model parameters are trained using the Adam optimizer \citep{kingma2014adam} with an initial learning rate of $2 \times 10^{-4}$ which is linearly decayed.
For span representation, we use the embedding function described in \citet{lee-etal-2017-end}.
For OntoNotes we follow the setup of \citet{xia2020revisiting}. We differ, however, in training all the model parameters, except SpanBERT, from scratch.
The models are trained for a maximum of 15 epochs for OntoNotes, and 25 epochs for LitBank. For both the datasets, the training stops if dev performance doesn't improve for 5 epochs. For more details see Appendix~\ref{sec:app_hyperparam}.

\section{Results}

\begin{table}[t]
\centering{
\small{
\caption{Results for OntoNotes (CoNLL F1) .}
\label{tab:res_ontonotes}

\begin{tabular}{lcc}
\toprule
Model & Dev F1 & Test F1 \\\midrule
\unbounded & 78.4 & 78.1  \\
U-MEM* & 79.6 & 79.6 \\
\learned &  & \\
\hspace{0.1in} 5 cells  & 74.0 & 73.3  \\
\hspace{0.1in} 10 cells  & 77.1 & 76.8  \\
\hspace{0.1in} 20 cells  & 78.1 & 78.2  \\
\lru & & \\
\hspace{0.1in} 5 cells   & 69.8 & 69.6  \\
\hspace{0.1in} 10 cells   & 75.9 & 75.5  \\
\hspace{0.1in} 20 cells   & 78.2 & 77.8  \\\midrule
U-MEM* (\citealp{xia2020revisiting}) & 79.7 & 79.4 \\\midrule
\citet{joshi2020span} & 80.1 & 79.6 \\
\citet{wu2019coreference} & 83.4 & 83.1 \\
\bottomrule
\end{tabular}
}
}
\end{table}

Tables \ref{tab:res_litbank} and~\ref{tab:res_ontonotes} show results of all the proposed models for LitBank and OntoNotes respectively. Detailed results with the performance on different coreference metrics is presented in Table~\ref{tab:full_litbank} for LitBank, and Table~\ref{tab:full_ontonotes} for OntoNotes (Appendix~\ref{sec:misc}).

As expected, the bounded memory models improve with increase in memory.
For both the datasets, the \learned model with 20 memory cells is competitive with the \unbounded model though the gap between them for LitBank is non-trivial. 
Among the bounded memory models, the \learned model is significantly better than \lru for lower numbers of memory cells. We analyze the reasons for this in the next section.

For OntoNotes, the U-MEM* model easily outperforms the \unbounded model which is trained to ignore all non-gold mentions.
These non-gold mentions also include singletons in case of OntoNotes. 
Thus, the  \unbounded model essentially has to predict if a non-coreferent mention will be coreferent with future mentions or not. 
U-MEM* avoids this difficult problem by adding all non-coreferent mentions to memory.
Since in OntoNotes singletons are removed during evaluation, the U-MEM* model is not penalized for predicting singletons corresponding to invalid mentions, and otherwise.
Note that the U-MEM* model doesn't make sense for LitBank where singletons are used for evaluation. The initial empirical results also confirmed that decisively.

Between the two datasets, we see that the increase in memory results in larger improvement for LitBank.
We also establish a new state-of-the-art for LitBank with the \unbounded model.
For OntoNotes, the U-MEM* model matches the performance of a similar model by  \citet{xia2020revisiting}.
Remarkably, the two U-MEM* models almost match the performance of the computationally and memory intensive span-ranking model of \citet{joshi2020span} whose finetuned SpanBERT document encoder is used by these two models.  
We expect gains by further finetuning the SpanBERT model and learning a parameterized global entity representation, but we leave them for future work.

\section{Analysis}
In this section we analyze the behavior of the three memory models on LitBank and OntoNotes.

\begin{table}[t!]

\centering{
\small{

\caption{Peak memory and inference time statistics for the LitBank cross-validation split $0$. Note that the training memory statistics depend on document truncation and sampling probability of invalid mentions. The models in this table are trained without document truncation and sample 20\% of invalid mentions during training.}%
\label{tab:mem_stats}
\begin{tabular}{l ccc}
\toprule
\multirow{2}{*}{Model} & Peak training   & Peak inference  & Inference  \\
     & mem. (in GB) &  mem. (in GB) &   time (in s)\\\midrule
\unbounded              &  11.6   &    3.1 &  29.25 \\
\learned    &      &                &            \\
\hspace{0.1in} 5 cells &  \phantom{1}8.0     &  3.2   &  27.31 \\
\hspace{0.1in} 10 cells &  \phantom{1}8.4    &  3.2   &  27.44 \\
\hspace{0.1in} 20 cells &  \phantom{1}9.1   &   3.2   &  27.86 \\
\lru & &  &  \\
\hspace{0.1in} 5 cells &  \phantom{1}8.0   &  3.2   & 26.19   \\
\hspace{0.1in} 10 cells &  \phantom{1}8.3   & 3.2   & 26.50  \\
\hspace{0.1in} 20 cells &  \phantom{1}8.9   &  3.2  & 26.19  \\\bottomrule
\end{tabular}
}
}
\end{table}

\begin{table}[t!]

\centering{
\small{

\caption{Comparison of number of entities in memory.}
\label{tab:num_ents}

\begin{tabular}{l c c c c}
\toprule
\multirow{2}{*}{Model} & \multicolumn{2}{c}{LitBank} & \multicolumn{2}{c}{OntoNotes} \\
     & Avg & Max & Avg & Max\\\midrule
\unbounded  & 80.8  & 160  & 16.0   &  \phantom{1}83 \\
U-MEM*  & -  & -  & 173   &  962 \\
\learned    &      &                &           & \\
\hspace{0.1in} 5 cells & \phantom{1}5.0  & \phantom{11}5   & \phantom{1}4.6       & \phantom{11}5 \\
\hspace{0.1in} 10 cells & 10.0 & \phantom{1}10 & \phantom{1}7.8 & \phantom{1}10 \\
\hspace{0.1in} 20 cells & 20.0 & \phantom{1}20 & 12.1      & \phantom{1}20 \\
\lru & &  & & \\
\hspace{0.1in} 5 cells & \phantom{1}5.0 & \phantom{11}5    & \phantom{1}4.6       & \phantom{11}5   \\
\hspace{0.1in} 10 cells & 10.0 & \phantom{1}10 & \phantom{1}7.9 & \phantom{1}10 \\
\hspace{0.1in} 20 cells & 20.0 & \phantom{1}20 & 11.9      & \phantom{1}20 \\\bottomrule
\end{tabular}
}
}

\end{table}

\paragraph{Memory Utilization}
Table~\ref{tab:mem_stats} compares the memory and inference time statistics for the different memory models for the LitBank cross-validation split zero.\footnote{Peak memory usage estimated via \texttt{torch.cuda.max\_memory\_allocated()}}
For training, the bounded memory models are significantly less memory intensive than the \unbounded model.
The table also shows that the bounded memory models are faster than the \unbounded memory model during inference (inference time calculated by averaging over three runs).
This is because the number of entities tracked by the \unbounded memory model grows well beyond the maximum of 20 memory slots reserved for the bounded models as shown in Table~\ref{tab:num_ents}.

Surprisingly, for inference we see that the bounded models have a slightly larger memory footprint than the \unbounded model.
This is because the document encoder, SpanBERT, dominates the memory usage during inference (as also observed by \citealp{xia2020revisiting}).
Thus the peak memory usage during inference is determined by the mention proposal stage rather than the mention clustering stage.
And during the mention proposal stage, the additional parameters of bounded memory models, which are loaded as part of the whole model, cause the slight uptick in peak inference memory.
Note that using a cheaper encoder or running on a sufficiently long document, such as a book,
can change these results.

\paragraph{Number of Entities in Memory}
Table~\ref{tab:num_ents} compares the maximum number of entities kept in memory by the different memory models for the LitBank cross-validation dev sets and the OntoNotes dev set.
As expected, the \unbounded model keeps more entities in memory than the bounded memory models on average for both datasets.
For LitBank the difference is especially stark with the  \unbounded model tracking about 4/8 times more entities in memory on average/worst case, respectively. 
The difference between the \unbounded and U-MEM* model is striking, with U-MEM* tracking more than 10 times the entities of \unbounded in both the average and worst case.
Also, while some OntoNotes documents do not use even the full 5 memory cell capacity, all LitBank documents fully utilize even the 20 memory cell capacity.
This is because %
LitBank documents are more than four times as long as OntoNotes documents, and LitBank has singletons marked. These results also justify our initial motivation that with long documents, the memory requirement will increase even if we only keep the entity representations.

\paragraph{\learned vs. \lru}

\begin{table}[t!]
\centering{
\small{
\caption{Average number of mentions ignored by the two bounded memory models.}
\label{tab:num_ignored}

\addtolength{\tabcolsep}{-3pt}
\begin{tabular}{c cccc}
\toprule
Memory & \multicolumn{2}{c}{LitBank} & \multicolumn{2}{c}{OntoNotes} \\
size     & \learned & \lru & \learned & \lru\\\midrule
\phantom{1}5 & \phantom{1}4.5  & 70.0   & 0.3       & 3.7 \\
10 & \phantom{1}0.0 & 14.2 & 0.0 & 0.4 \\
20 & \phantom{1}0.0 & \phantom{1}0.4 & 0.0  & 0.1 \\\bottomrule
\end{tabular}
}
}
\end{table}

Table~\ref{tab:num_ignored} compares the number of mentions ignored by \learned and \lru.
The \learned model ignores far fewer mentions than \lru.
This is because while the \lru model can only evict the LRU entity, which might not be optimal,
the \learned model can choose any entity for eviction.
These statistics combined with the fact that the \learned model typically outperforms \lru mean
that the \learned model is able to anticipate which entities are important and which are not.

\begin{table}[t!]

\centering{
\small{
\caption{Error Analysis for OntoNotes dev set. CE=Conflated Entities, DE=Divided Entity, EM=Extra Mention, EE=Extra Entity, MM=Missing Mention, ME=Missing Entity.}
\label{tab:err_analysis}
\begin{tabular}{l c c c c c c}
\toprule
Model &  CE & DE & EM & EE & MM & ME \\\midrule
\unbounded & 853& 496& 515& 904& 545& \phantom{1}603 \\
U-MEM* & 754& 466& 504& 816& 527& \phantom{1}583 \\
\learned    &      &     &      &    &   &  \\
\hspace{0.1in}  5 cells & 706& 381& 340& 972& 844& 1116 \\
\hspace{0.1in}  10 cells & 757& 425& 340& 868& 655& \phantom{1}894 \\
\hspace{0.1in}  20 cells &  752& 396& 402& 859& 613& \phantom{1}799 \\
\lru    &      &     &      &    &   & \\
\hspace{0.1in}  5 cells & 672& 369& 365& 1146& 923& 1359 \\
\hspace{0.1in}  10 cells & 722& 393& 403& 986& 696& \phantom{1}931 \\
\hspace{0.1in}  20 cells & 713& 420& 380& 833& 559& \phantom{1}853 \\\bottomrule
\end{tabular}
}
}

\end{table}

\paragraph{Error Analysis}
Table~\ref{tab:err_analysis} presents the results of automated error analysis done using the Berkeley Coreference Analyzer~\citep{kummerfeld2013error} for the OntoNotes dev set.
As the memory capacity of models increases, the errors shift from missing mention, missing entity, and divided entity categories, to conflated entities, extra mention, and extra entity categories.
The \learned model outperforms \lru in terms of tracking more entities. %

\section{Conclusion and Future Work}
We propose a memory model which tracks a small, bounded number of entities.
The proposed model guarantees a linear runtime in document length, and in practice significantly reduces peak memory usage during training.
Empirical results on LitBank and OntoNotes show that the model is competitive with an unbounded memory version and outperforms a strong rule-based baseline.
In particular, we report state of the art results on LitBank.
In future work we plan to apply our model to longer, book length documents, and plan to add more structure to the memory.

\section*{Acknowledgments}

We thank David Bamman for help with the LitBank setup, and Patrick Xia for answering questions about their coreference model. We also thank the anonymous EMNLP reviewers for their valuable feedback.
This material is based upon work supported by the National Science Foundation under Award Nos.~1941178 and 1941160.

\begin{figure*}[h]
\centering
\begin{subfigure}[b]{0.45\textwidth}
        \centering
        \includegraphics[width=\textwidth]{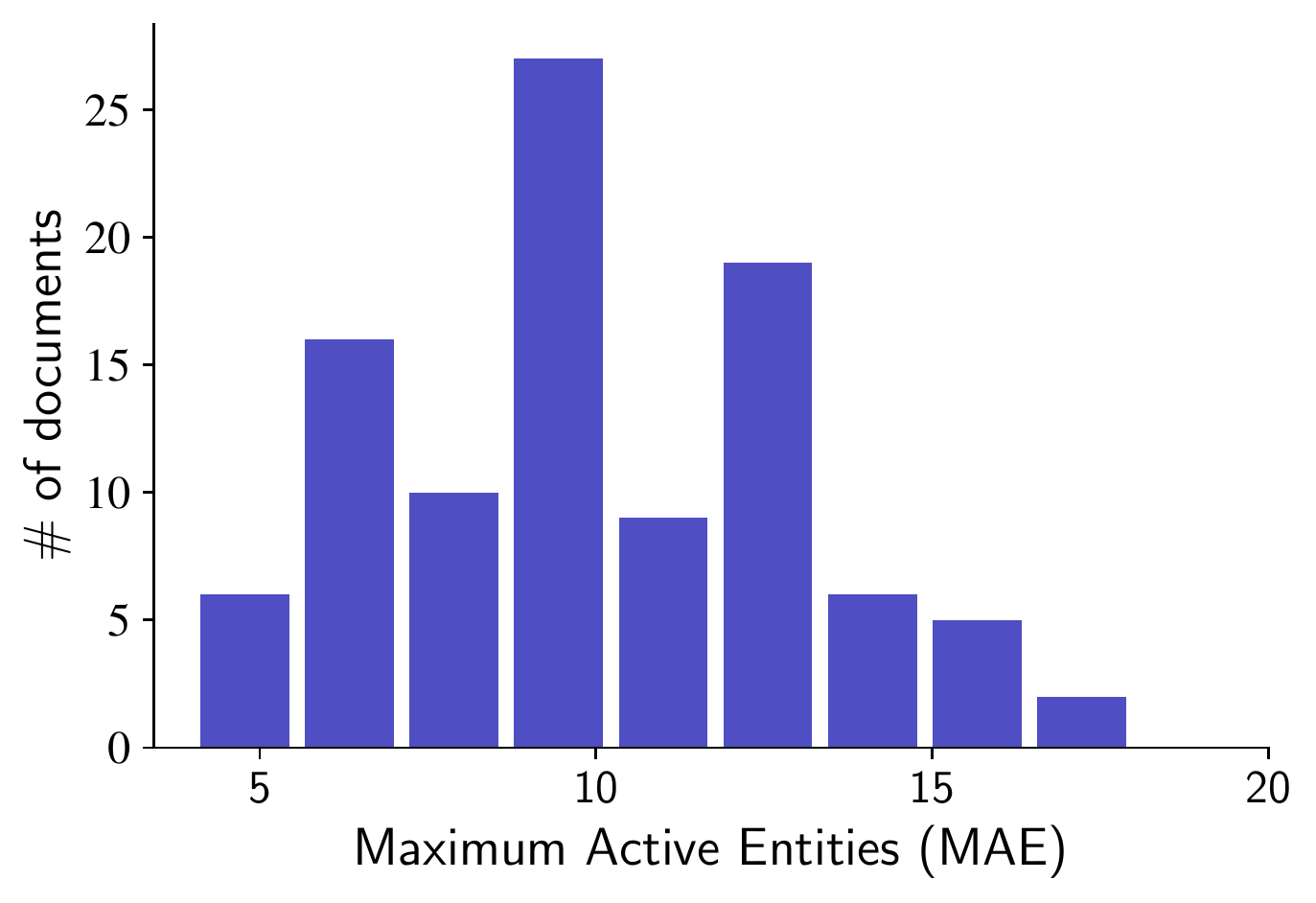}
        \caption{LitBank.}
        \label{fig:litbank_active}
    \end{subfigure}%
    ~
    \begin{subfigure}[b]{0.45\textwidth}
        \centering
        \includegraphics[width=\textwidth]{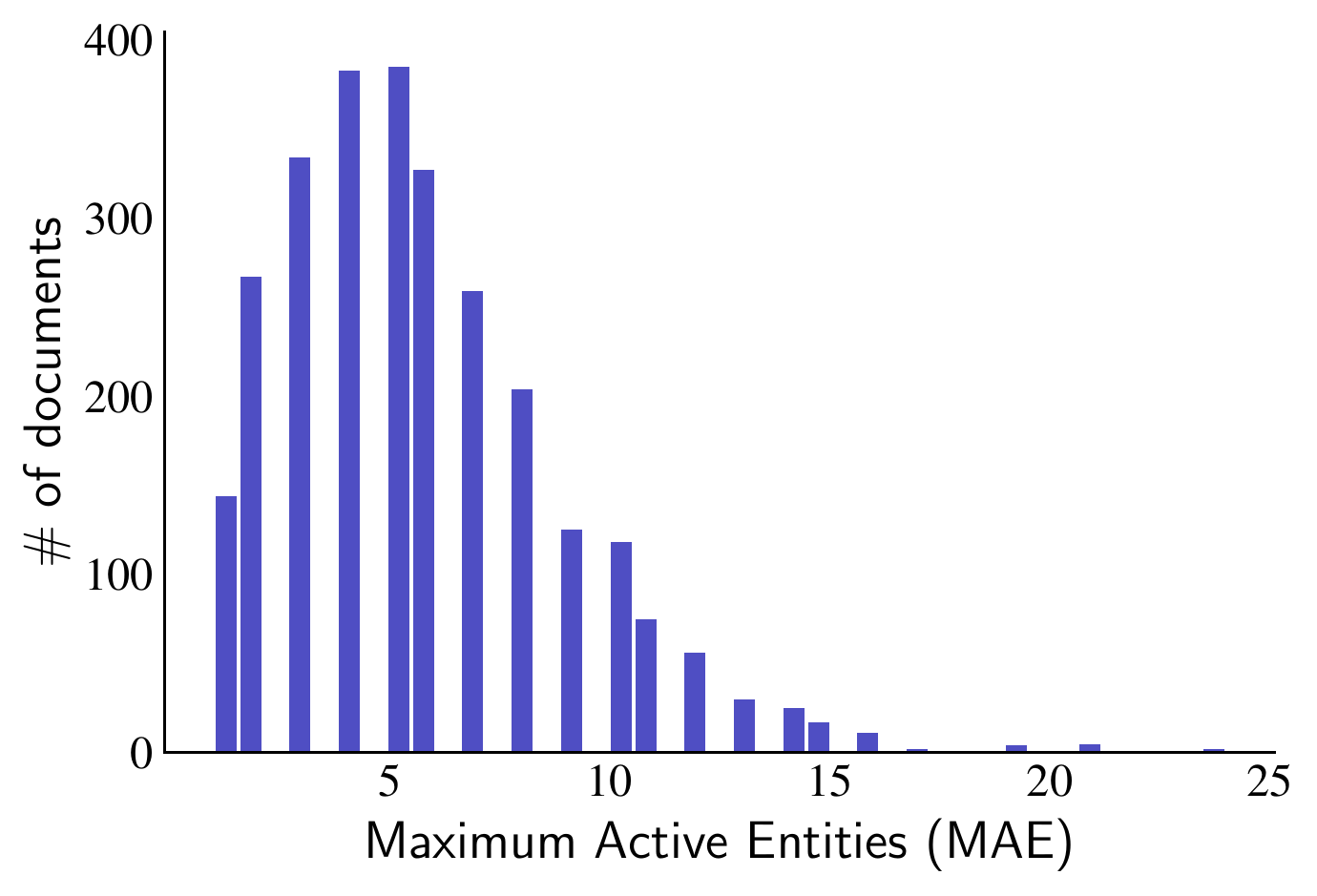}
        \caption{OntoNotes Training Set.}
        \label{fig:ontonotes_active}
    \end{subfigure}
    \caption{Histograms of Maximum Active Entities for documents in LitBank and OntoNotes.}
    \label{fig:active_chains}
\end{figure*}

\begin{figure*}[h]
\centering
\begin{subfigure}[b]{0.45\textwidth}
        \centering
        \includegraphics[width=\textwidth]{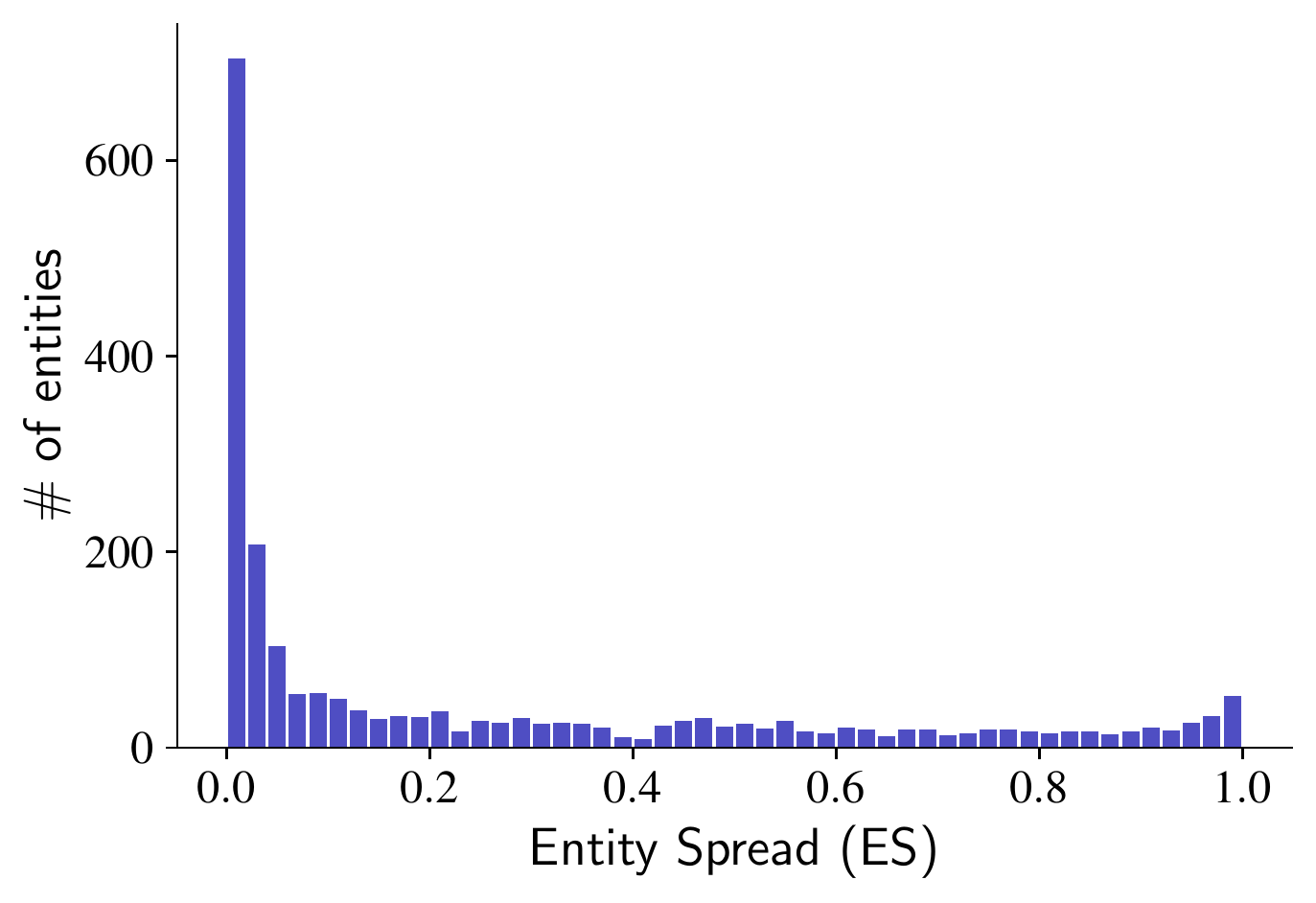}
        \caption{LitBank.}
        \label{fig:litbank_spread}
    \end{subfigure}%
    ~
    \begin{subfigure}[b]{0.45\textwidth}
        \centering
        \includegraphics[width=\textwidth]{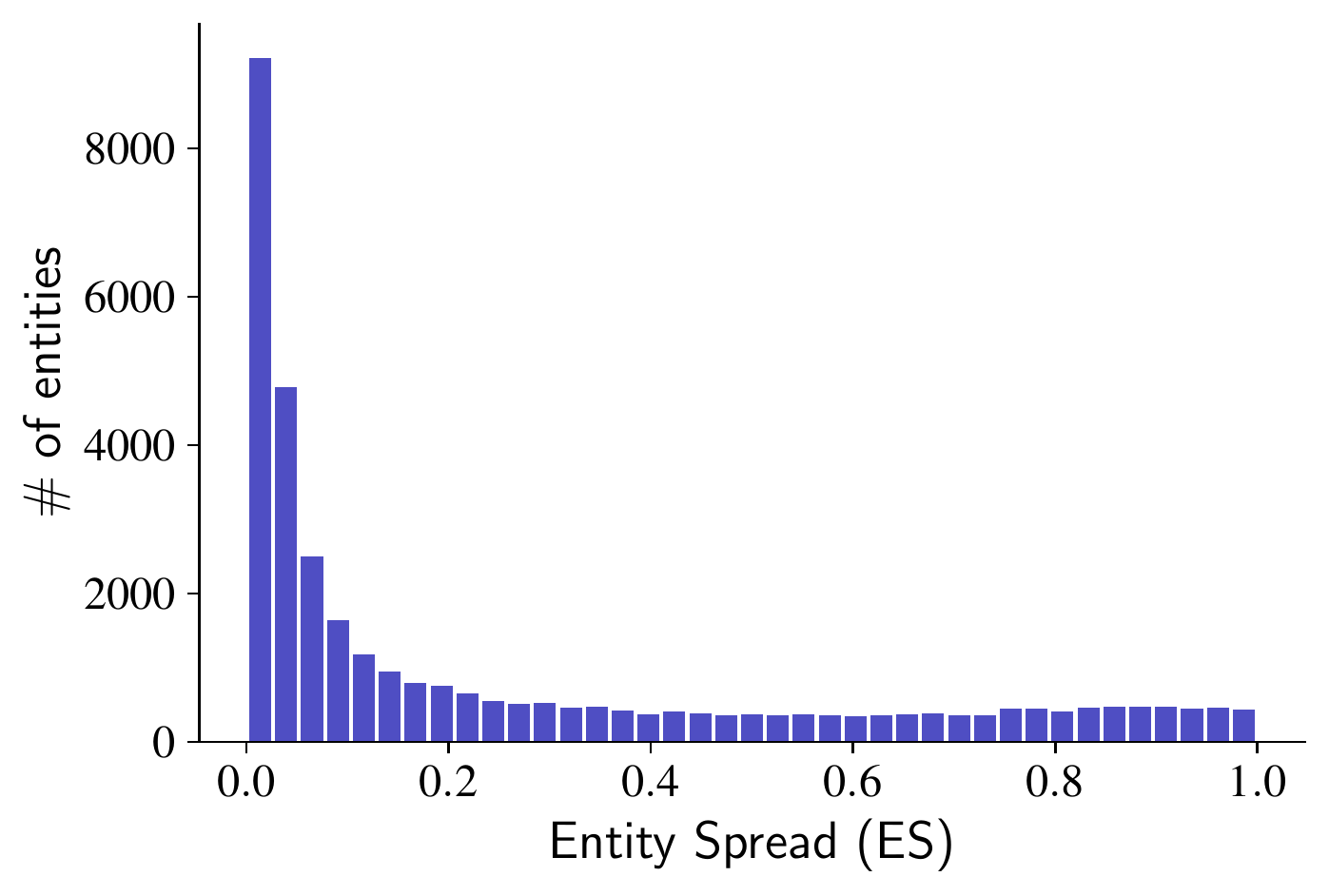}
        \caption{OntoNotes Training Set.}
        \label{fig:ontonotes_spread}
    \end{subfigure}
    \caption{Histograms of Entity Spread as fraction of document length for LitBank and OntoNotes.}
    \label{fig:entity_spread}
\end{figure*}

\bibliography{0-main}
\bibliographystyle{acl_natbib}
\newpage

\appendix
\section{Appendix}

\subsection{Maximum Active Entities}
Figure~\ref{fig:active_chains} visualizes the histograms of length of Entity Spread (ES), defined in Section~\ref{sec:active},  as a fraction of document length for documents in LitBank and OntoNotes. For LitBank we only visualize the entity spread of non-singleton clusters because otherwise the histogram is too skewed towards one.
Figure~\ref{fig:entity_spread} visualizes the histograms of Maximum Active Entity Count (MAE), defined in Section~\ref{sec:active}, for documents in LitBank and OntoNotes.

\begin{table}[!ht]

\centering{
\small{
\caption{Hyperparameter options for OntoNotes with preferred choices highlighted in bold.}
\label{tab:hyperparam_ontonotes}
\begin{tabular}{ll}
\toprule
Parameter & Range \\\midrule
Dropout & \{0.4\}\\
FFNN hidden layer & \{3000\}\\
FFNN \# of hidden layers & 1\\
Document Encoding & \{Independent, \textbf{Overlap}\} \\
Label Smoothing & \{0.0, 0.01, 0.1\} \\
Sampling Prob. Invalid Mentions & \{0.25, 0.5, 0.75, 1.0\} \\
Max. \# of BERT Segments & \{\textbf{3}, 5\} \\
Non-coreferent entity weight & \{1.0\}\\
\bottomrule
\end{tabular}
}
}
\end{table}

\subsection{Model Details}
\label{sec:app_hyperparam}

\paragraph{Hyperparameter Choices}
We stick with the hyperparameters for feedforward neural network (FFNN) size and depth from~\citet{joshi2020span}. We didn't do much exploration with dropout but with the limited experiments our finding was that there was little separating dropout probability of 0.3 and 0.4. Among choices for how to segment document into BERT windows, we found overlapping windows to work better than independent BERT windows. Two very important hyperparameters that affect peak memory usage during training are: (a) maximum number of BERT segments, and (b) sampling probability of invalid mentions. Truncating the document by selecting a chosen maximum number of contiguous BERT segments essentially caps the length of documents during training. And the second hyperparameter of sampling invalid mentions  controls the number of invalid mentions, which happens to be the overwhelming category of proposed mentions, the model sees during training. We also explore two hyperparameters for the cross-entropy loss of the first step of mention clustering: (a) label smoothing for regularization, and (b) weight of the non-coreferent term in the cross-entropy loss.

\paragraph{Specific Hyperparameter Choices for OntoNotes}
We didn't see any gain by increasing the maximum number of BERT segments from 3 to 5 in our initial experiments. The \unbounded and bounded models preferred lower sampling probabilities for invalid mentions but no clear winner in label smoothing weight. The U-MEM* model preferred low label smoothing weight and higher sampling probabilities for invalid mentions.

\paragraph{Specific Hyperparameter Choices for LitBank}
Initial experiments with cross validation splits \{0, 1, 2\} showed that models preferred maximum number of BERT segments to be 5 in comparison to 3. This might be because most of the LitBank documents are really long, and training on a maximum of 3 BERT segments might lead to a bigger mismatch between training and inference. 
Another hyperparameter that proved important for LitBank was the non-coreferent entity weight of $2.0$. Due to explosion of combinations driven by the fact that there are 10 cross validation splits, we didn't explore label smoothing for LitBank.

\begin{table}[t]

\centering{
\small{
\caption{Hyperparameter options for LitBank with  preferred choices highlighted in bold.}
\label{tab:hyperparam_litbank}
\begin{tabular}{ll}
\toprule
Parameter & Range \\\midrule
Dropout & \{0.3\}\\
FFNN hidden layer & \{3000\}\\
FFNN \# of hidden layers & 1\\
Document Encoding & \{Independent, \textbf{Overlap}\} \\
Label Smoothing & \{0.0\} \\
Sampling Prob. Invalid Mentions & \{0.25, 0.5, 0.75, 1.0\} \\
Max. \# of BERT Segments & \{3, \textbf{5}\} \\
Non-coreferent entity weight & \{1.0, \textbf{2.0}\}\\
\bottomrule
\end{tabular}
}
}
\end{table}

\subsection{Ground Truth Generation}
\label{sec:app_gt}
In this section we explain how the ground truth action sequence is generated corresponding to the predicted mention sequence.
The ground truth for \unbounded model is fairly straight forward.
For the bounded memory models, we keep growing the number of entities till we hit the memory ceiling.
For all the entities in memory, we maintain the number of mentions remaining in the ground truth cluster.
For example, a cluster with a total of five mentions, two of which have already been processed by the model, has three remaining mentions.

Suppose now a mention corresponding to a currently untracked entity comes in and the memory is already at full capacity.
Then for the \learned model, we compare the number of mentions of this new entity (along with the current mention) against the number of mentions remaining for all the entities currently being tracked. If there are entities in memory with number of remaining mentions less than or equal to the number of mentions of this currently untracked entity, then the untracked entity replaces the entity with the least number of remaining mentions. Ties among the entities with least number of remaining mentions are broken by the least recently seen entity. If there's no such entity in the memory, then the mention is ignored.
For the \lru model, the comparison is done in a similar way but is limited to just the LRU entity.

\begin{table}[!t]
    \centering{
    \small{

    \caption{Number of model parameters (in millions).}
    \label{tab:num_param}

    \begin{tabular}{lll}
        \toprule
                & LitBank & OntoNotes \\\midrule
    \unbounded  & 37.36 & 37.42\\
    \learned    & 46.83 & 46.95\\
    \lru        & 46.83 & 46.95\\\bottomrule
\end{tabular}
}
}
\end{table}

\subsection{Miscellany}
\label{sec:misc}
\begin{table*}[!ht]

\centering{
\caption{Detailed results of the proposed models on the aggregated LitBank cross-validation test set.}
\label{tab:full_litbank}

\begin{tabular}{l c c c c c c c c c c}
\toprule
Model & \multicolumn{3}{c}{MUC} & \multicolumn{3}{c}{$\text{B}^3$} & \multicolumn{3}{c}{$\text{CEAF}_{\phi_4}$} & \\
 & Prec. & Rec. & F1 & Prec. & Rec. & F1 & Prec. & Rec. & F1 & Avg.\ F1\\\midrule
 \unbounded & 90.8 & 85.7 & 88.2  & 80.0 & 72.1 & 75.9  & 65.1 & 66.0 & 65.5 &  76.5\\
\learned \\
\hspace{0.1in} 5 cells & 90.9 & 80.0 & 85.1  & 77.4 & 64.0 & 70.1  & 57.8 & 53.8 & 55.7 &  70.3\\
\hspace{0.1in} 10 cells & 90.0 & 84.6 & 87.2  & 78.1 & 70.8 & 74.2  & 64.2 & 61.1 & 62.6 &  74.7\\
\hspace{0.1in} 20 cells & 90.3 & 85.0 & 87.6  & 79.2 & 70.9 & 74.8  & 64.1 & 62.0 & 63.0 &  75.1\\
\lru \\
\hspace{0.1in} 5 cells & 91.4 & 74.6 & 82.2  & 75.7 & 51.1 & 61.0  & 52.2 & 21.3 & 30.3 &  57.8\\
\hspace{0.1in} 10 cells & 91.1 & 81.3 & 85.9  & 78.5 & 62.1 & 69.3  & 56.3 & 47.8 & 51.7 &  69.0\\
\hspace{0.1in} 20 cells & 90.5 & 85.1 & 87.7  & 79.7 & 69.8 & 74.4  & 61.1 & 61.0 & 61.1 &  74.4\\
\bottomrule
\end{tabular}
}
\end{table*}

\begin{table*}[t]

\centering{
\caption{Detailed results of the proposed models on the OntoNotes 5.0 test set.}
\label{tab:full_ontonotes}
\begin{tabular}{l c c c c c c c c c c}
\toprule
Model & \multicolumn{3}{c}{MUC} & \multicolumn{3}{c}{$\text{B}^3$} & \multicolumn{3}{c}{$\text{CEAF}_{\phi_4}$} & \\
 & Prec. & Rec. & F1 & Prec. & Rec. & F1 & Prec. & Rec. & F1 & Avg.\ F1\\\midrule
\unbounded &  84.6 &  84.1 &  84.3  &  77.2 &  76.2 &  76.7  &  72.5 &  74.3 &  73.4  & 78.1  \\
U-MEM* &  85.5 & 85.1 &  85.3 & 78.7 & 77.3 & 78.0  &  74.2 & 76.5 &  75.3  & 79.6  \\

\learned   \\
\hspace{0.1in} 5 cells &  76.4 &  86.2 &  81.0  &  66.4 &  78.4 &  71.9  &  62.0 &  72.7 &  66.9  & 73.3  \\
\hspace{0.1in} 10 cells &  81.7 &  85.9 &  83.8  &  72.8 &  77.9 &  75.3  &  67.0 &  76.4 &  71.4  & 76.8  \\
\hspace{0.1in} 20 cells &  83.2 &  86.2 &  84.7  &  74.8 &  78.9 &  76.8  &  70.0 &  76.7 &  73.2  & 78.2  \\
\hspace{0.1in} 30 cells &  83.8 &  85.6 &  84.7  &  76.1 &  78.2 &  77.1  &  70.4 &  77.1 &  73.6  & 78.5  \\
\lru \\

\hspace{0.1in} 5 cells  &  72.0 &  85.7 &  78.3  &  60.1 &  78.9 &  68.2  &  57.0 &  68.9 &  62.4  & 69.6  \\
\hspace{0.1in} 10 cells  &  80.1 &  85.7 &  82.8  &  70.5 &  78.3 &  74.2  &  66.0 &  73.4 &  69.5  & 75.5  \\
\hspace{0.1in} 20 cells  &  82.8 &  85.9 &  84.3  &  74.8 &  78.3 &  76.5  &  68.0 &  77.4 &  72.4  & 77.8  \\
\hspace{0.1in} 30 cells  &  84.0 &  85.2 &  84.6  &  76.2 &  78.2 &  77.2  &  72.1 &  75.6 &  73.8  & 78.5  \\
\bottomrule
\end{tabular}
}
\end{table*}

\paragraph{Computing Infrastructure \& Runtime}
All the models for a single cross validation split of LitBank can be trained within 4 hours. Training on OntoNotes finishes within 12-20 hours.
The U-MEM* model where all invalid mentions are seen during training is the only configuration that requires 24GB memory GPUs, all other configurations can be trained on 12GB memory GPUs.

\paragraph{Number of model parameters.}

Table~\ref{tab:num_param} shows the number of trainable parameters for all the model and dataset combinations. \learned and \lru have additional parameters in comparison to \unbounded for predicting a score corresponding to the number of remaining mentions for an entity.
Comparing across datasets, the OntoNotes models  have a few additional parameters than their LitBank counterparts for modeling the document genre.

\paragraph{Evaluation Metric Code.}
We use the coreference scorer Perl script available at \url{https://github.com/conll/reference-coreference-scorers}.
We also use the Python implementation by Kenton Lee available at \url{https://github.com/kentonl/e2e-coref/blob/master/metrics.py}.
The two scripts can have some rounding differences.

\begin{table}[t!]

\centering{
\small{
\caption{Spearman correlation of F1 score with document length and \# of entities in OntoNotes dev set.}
\label{tab:corr}

\begin{tabular}{l c c }
\toprule
Model & Document Length & \# of Entities \\\midrule
\unbounded  & -0.31  & -0.28 \\
U-MEM* & -0.28 & -0.25 \\
\learned   &          & \\
\hspace{0.1in} 5 cells & -0.36 & -0.37 \\
\hspace{0.1in} 10 cells & -0.34 & -0.33 \\
\hspace{0.1in} 20 cells & -0.34 & -0.31 \\

\lru & &  \\
\hspace{0.1in} 5 cells & -0.37 & -0.41   \\
\hspace{0.1in} 10 cells & -0.29 & -0.30 \\
\hspace{0.1in} 20 cells& -0.31 & -0.29 \\\bottomrule
\end{tabular}
}
}
\end{table}
\paragraph{Effect of Document Length and Number of Entities.}
Table~\ref{tab:corr} and ~\ref{tab:corr_litbank} presents the Spearman correlation of document F1 score with  %
document length and number of entities in the document.
The correlations are mostly negative because the task becomes more challenging with an increase in document length and entities, though for LitBank the length of the document  doesn't seem to be a great indicator of the hardness of the task. 
The increase in memory capacity for bounded models results in less negative correlations, suggesting improved performance for challenging documents.

\begin{table}[t!]

\centering{
\small{
\caption{Spearman correlation of F1 score with document length and \# of entities in aggregated LitBank cross-validation dev set.}
\vspace{0.00in}
\label{tab:corr_litbank}
\begin{tabular}{l c c }
\toprule
Model & Document Length & \# of Entities \\\midrule
\unbounded  & -0.07 & -0.36 \\
\learned   &          & \\
\hspace{0.1in} 5 cells & \phantom{-}0.01 & -0.23 \\
\hspace{0.1in} 10 cells & -0.06 & -0.35 \\
\hspace{0.1in} 20 cells & -0.03 & -0.32 \\

\lru & &  \\
\hspace{0.1in} 5 cells & -0.00 & -0.41  \\
\hspace{0.1in} 10 cells & \phantom{-}0.01 & -0.36 \\
\hspace{0.1in} 20 cells& -0.02 & -0.33 \\\bottomrule
\end{tabular}
}
}
\end{table}

\end{document}